\documentclass{article}
\usepackage{amsmath}
\usepackage{graphicx}
\usepackage{float}

\usepackage[dblblindworkshop, final]{neurips_2025}
 \workshoptitle{Evaluating the Evolving LLM Lifecycle: Benchmarks, Emergent Abilities, and Scaling \textbf{[selected for contributed talk]}}

\usepackage[utf8]{inputenc} % allow utf-8 input
\usepackage[T1]{fontenc}    % use 8-bit T1 fonts
\usepackage{hyperref}       % hyperlinks
\usepackage{url}            % simple URL typesetting
\usepackage{booktabs}       % professional-quality tables
\usepackage{amsfonts}       % blackboard math symbols
\usepackage{nicefrac}       % compact symbols for 1/2, etc.
\usepackage{microtype}      % microtypography
\usepackage{xcolor}         % colors

\title{LLMs Show Surface-Form Brittleness Under Paraphrase Stress Tests}

\author{%
    Juan Miguel Navarro Carranza \\
    Stanford University\\
  \texttt{jmnavarr@stanford.edu} \\
}

\begin{document}

\maketitle

\begin{abstract}
Benchmark scores for Large Language Models (LLMs) can be inflated by \emph{memorization} of test items or near duplicates. We present a simple, protocol that probes \emph{generalization} by re-evaluating models on \emph{paraphrased} versions of benchmark questions. Using Mistral-7B-Instruct and Qwen2.5-7B-Instruct, we measure the accuracy gap between original and paraphrased items on ARC-Easy and ARC-Challenge. Our pipeline controls decoding, enforces multiple-choice output format, and includes a robust paraphrase-cleaning step to preserve semantics. We find that paraphrasing induces a non-trivial accuracy drop (original vs. paraphrased), consistent with prior concerns about contamination and brittle surface-form shortcuts. 
\end{abstract}

%========================
% Result Macros (fill these)
%========================

\newcommand{\NArcEasy}{\textbf{300}}        % e.g., 300
\newcommand{\NArcChal}{\textbf{299}}        % e.g., 300

\newcommand{\AccEasyOrig}{\textbf{0.90}}      % e.g., 0.78
\newcommand{\AccEasyPara}{\textbf{0.84}}      % e.g., 0.70
\newcommand{\DropEasy}{\textbf{0.06}}              % = AccEasyOrig - AccEasyPara
\newcommand{\AccChalOrig}{\textbf{0.89}}      % e.g., 0.55
\newcommand{\AccChalPara}{\textbf{0.83}}      % e.g., 0.45
\newcommand{\DropChal}{\textbf{0.07}}              % = AccChalOrig - AccChalPara

\newcommand{\AccEasyOrigSec}{\textbf{0.84}}      % e.g., 0.78
\newcommand{\AccEasyParaSec}{\textbf{0.74}}      % e.g., 0.70
\newcommand{\DropEasySec}{\textbf{0.10}}              % = AccEasyOrig - AccEasyPara
\newcommand{\AccChalOrigSec}{\textbf{0.75}}      % e.g., 0.55
\newcommand{\AccChalParaSec}{\textbf{0.69}}      % e.g., 0.45
\newcommand{\DropChalSec}{\textbf{0.06}}              % = AccChalOrig - AccChalPara

\newcommand{\Mistral}{Mistral-7B-Instruct}    
\newcommand{\Qwen}{Qwen2.5-7B-Instruct} 
%========================
\section{Introduction}
Recent work shows that LLMs can regurgitate training data~\citep{carlini_extracting_2021}, and that duplication in pretraining corpora amplifies this effect~\citep{kandpal_deduplicating_2022}. As a result, static benchmarks may overestimate model capability, especially when test items or near duplicates leak into training sets~\citep{dong_generalization_2024}. A complementary perspective focuses on \emph{behavioral robustness} to surface perturbations---if a model relies on phrasing, small paraphrases can cause large performance swings~\citep{ribeiro_beyond_2020}. Code-evaluation work reinforces this: evolved or mutated prompts can sharply reduce apparent competence~\citep{liu_is_2023}.

\paragraph{Contributions.}
(1) A protocol to measure generalization via paraphrase stress tests; (2) a paraphrase–cleaning pipeline that preserves semantics while removing formatting artifacts; (3) a fully specified setup (\emph{Benchmark:} ARC-Easy/Challenge~\citep{clark_think_2018}, \emph{Models:} Mistral-7B-Instruct, Qwen2.5-7B-Instruct for answering and paraphrasing) using 4-bit inference on a single A100; (4) empirical findings that paraphrasing induces a measurable accuracy drop, suggesting residual memorization/brittleness; (5) released code for reproducibility.

%========================
\section{Related Work}
\textbf{Memorization and privacy.} Training-data extraction and memorization in LMs are well documented~\citep{carlini_extracting_2021}; deduplication mitigates leakage and privacy risk~\citep{kandpal_deduplicating_2022}. 
\textbf{Contamination and trustworthy eval.} Methods to detect contamination and separate memorization from generalization continue to emerge~\citep{dong_generalization_2024}. 
\textbf{Behavioral robustness.} CheckList formalizes capability/behavior tests with perturbations~\citep{ribeiro_beyond_2020}. 
\textbf{Evolved tests.} For code, augmented or mutated tests (e.g., EvalPlus) expose spurious success~\citep{liu_is_2023}. We adopt a minimal paraphrase-only variant that applies across QA benchmarks.

%========================

\section{Method}
\label{sec:method}

We assess surface–form robustness with a \emph{paraphrase stress test}. We use \emph{surface–form robustness} to mean that, under meaning–preserving rewordings (paraphrases, light syntactic changes, formatting tweaks), a model’s predictions remain unchanged. For each multiple–choice item, we evaluate an instruction–tuned LLM once on the original question and once on a semantically equivalent paraphrase, then report the paired accuracy gap $\Delta$ between the two conditions.

\subsection{Task and Data}
We consider multiple–choice question answering (MCQ). Each item $i$ comprises a question $q_i$, an option set $C_i=\{c_{i1},\dots,c_{ik_i}\}$, and a ground–truth answer letter $a_i\in\{A,\dots\}$. Our experiments use the ARC benchmark~\citep{clark_think_2018} (Hugging Face distribution~\citep{allen_ai_allenaiai2_arc_2025}), evaluating both \textsc{ARC–Easy} and \textsc{ARC–Challenge} on their validation splits. We evaluate a fixed subset (ARC–Easy: $\NArcEasy$, ARC–Challenge: $\NArcChal$) selected deterministically with a fixed random seed. Only questions are paraphrased; answer options $C_i$ are kept verbatim to preserve label mappings.

\subsection{Models and Inference}
\label{sec:models}

\paragraph{Roles and cross–pairing.}
We use two instruction–tuned decoder--only LLMs and assign disjoint roles:
\emph{Answerer} (produces the MCQ choice) and \emph{Paraphraser} (rewrites only the question stem).
To discourage style echo and self–consistency artifacts, the paraphraser is always a \emph{different model family} than the answerer.

We evaluate \texttt{mistralai/Mistral-7B-Instruct-v0.3}~\citep{mistral_ai_mistralaimistral-7b-instruct-v03_2025,jiang_mistral_2023} and \texttt{Qwen2.5-7B-Instruct}~\citep{qwen_team_qwenqwen25-7b-instruct_2025,yang_qwen2_2024}. The paraphraser always uses the model not acting as the answerer in that run (cross–family separation). We report both cross–pairings:
(\Mistral\ $\rightarrow$ answerer, \Qwen\ $\rightarrow$ paraphraser) and
(\Qwen\ $\rightarrow$ answerer, \Mistral\ $\rightarrow$ paraphraser).

\paragraph{Compute and environment.}
All runs are executed on a single NVIDIA A100--SXM4--80GB (CUDA~12.4, driver~550.54.15).
Models are loaded via \texttt{transformers} with 4--bit quantization (\texttt{bitsandbytes} NF4) using \texttt{device\_map="auto"}.
Randomness is controlled via \texttt{random.seed(1337)} and \texttt{torch.manual\_seed(1337)}.
No few–shot examples, chain–of–thought, tools, or retrieval are used.

\paragraph{Quantization.}
We follow a standard 4--bit inference recipe:
\texttt{load\_in\_4bit=True}, \texttt{bnb\_4bit\_use\_double\_quant=True},
\texttt{bnb\_4bit\_quant\_type="nf4"}, and
\texttt{bnb\_4bit\_compute\_dtype=torch.bfloat16}.
This yields stable, memory–efficient inference on the A100 while preserving accuracy in our setting.

\paragraph{Prompts.}
\emph{Answering.} We format each item with the question followed by lettered options A, B, C, \dots\ and require a \emph{single–letter} decision.
When \texttt{force\_letter=True}, the instruction requires a one–object JSON:
\texttt{\{"answer": "LETTER", "explanation": "1--2 sentences"\}},
and forbids any extra text.

\emph{Paraphrasing.} We use a constrained rewrite prompt that (i) rewrites the \emph{question stem} only, (ii) preserves all meaning and details \emph{verbatim} (numbers/units/entities), (iii) outputs only the rewritten stem without labels or commentary.

\paragraph{Decoding.}
\emph{Answering (deterministic):} \texttt{max\_new\_tokens}=32, \texttt{temperature}=0.0, \texttt{top\_p}=1.0, \texttt{do\_sample=False}.
Determinism ensures fair comparisons across original vs.\ paraphrased conditions.

\emph{Paraphrasing (light sampling):} \texttt{max\_new\_tokens}=128, \texttt{temperature}=0.7, \texttt{top\_p}=0.95, \texttt{do\_sample=True}.
Mild diversity avoids trivial restatements while remaining close to the source; fidelity checks (below) prevent semantic drift.

\paragraph{Output parsing and formatting guardrails.}
For the answerer, we parse the JSON field \texttt{answer} when present and validate it against the option set.
If the model emits residual text, we fall back to a strict letter extractor that (i) prefers the JSON key when available, (ii) otherwise searches for a leading pattern like \texttt{"Answer:\ <LETTER>"} and (iii) rejects spurious matches (e.g., picking the first uppercase letter encountered).
This avoids the common failure mode where generic capital letters (e.g., “A” at the start of “Answer: C”) are misread as the choice.
We score with exact letter agreement against the gold key.

\paragraph{Paraphrase fidelity checks and retries.}
We paraphrase the stem once and run a sequence of automatic checks; if a check fails, we \emph{retry} up to $K{=}3$ times with the same decoding settings:
(i) nonempty output after cleaning (trim quotes/bullets, strip prefixes);
(ii) output differs from the original stem case–insensitively;
(iii) minimum length threshold: $\mathrm{len}(\tilde{q}) \ge \max(10,\ 0.6\,\mathrm{len}(q))$ to avoid fragmentary rewrites.
Only stems that pass all checks are accepted; otherwise we fall back to the original stem to keep the item count fixed.
Answer options are \emph{never} paraphrased to preserve label mappings.

\paragraph{Dataset slice and evaluation protocol.}
Unless stated otherwise, we evaluate ARC validation splits with a fixed, deterministic subset size (e.g., \NArcEasy\ for \textsc{ARC-Easy}, \NArcChal\ for \textsc{ARC-Challenge}). For each item, let $r_i^{\text{orig}}$ and $r_i^{\text{para}}$ be predictions on original and paraphrased questions, respectively. Define correctness
$o_i=\mathbb{I}[r_i^{\text{orig}}{=}a_i]$ and $p_i=\mathbb{I}[r_i^{\text{para}}{=}a_i]$.
We report
\[
\text{Acc}_{\text{orig}}=\frac{1}{n}\sum_{i=1}^{n} o_i,\quad
\text{Acc}_{\text{para}}=\frac{1}{n}\sum_{i=1}^{n} p_i,\quad
\Delta=\text{Acc}_{\text{orig}}-\text{Acc}_{\text{para}}.
\]

%========================

\section{Results}
\label{sec:results}

\paragraph{Overall trends.} Across both ARC splits, paraphrasing consistently reduces accuracy ($\Delta > 0$).
Table~\ref{tab:main} shows drops ranging from $0.06$ to $0.10$, confirming that
surface--form changes measurably degrade performance even when semantic content is preserved.
The effect is robust across cross--pairings: regardless of whether \Mistral\ or \Qwen\ is the answerer,
accuracy on paraphrased items is lower than on the original items.

\paragraph{Dataset difficulty.} Absolute accuracy is higher on \textsc{ARC-Easy} than on \textsc{ARC-Challenge} for both models,
consistent with the benchmark design. However, relative brittleness is not uniform:
the largest drop occurs for \Mistral\ answering \textsc{ARC-Easy} ($\Delta = 0.10$),
while both models show more moderate drops on \textsc{ARC-Challenge} ($\Delta = 0.06$--$0.07$).
This suggests that even “easier” items are fragile under rewording.

\paragraph{Cross--model comparison.} When acting as the answerer, \Qwen\ achieves higher baseline accuracy
($0.90$ on Easy, $0.89$ on Challenge) than \Mistral\ ($0.84$ and $0.75$ respectively).
Yet both exhibit similar paraphrase sensitivity ($\Delta$ in the same $0.06$--$0.10$ band),
indicating that brittleness is not confined to a single model family but is a shared vulnerability.

\paragraph{Qualitative flips.}
We observe two categories of changes:
(i) \emph{original$\rightarrow$incorrect}, where a paraphrase induces an error despite a correct original response;
(ii) \emph{incorrect$\rightarrow$correct}, where paraphrasing helps the model recover the right answer.
The former dominates, but the latter occurs in a nontrivial minority of cases,
highlighting that paraphrasing does not simply act as uniform noise but can also reshape
decision boundaries in helpful ways.

\begin{table}[H]
\centering
\small
\caption{Accuracy on original vs.\ paraphrased items using Mistral-7B-Instruct and Qwen2.5-7B-Instruct. $\Delta$ quantifies brittleness to paraphrase.}
\resizebox{\textwidth}{!}{%
\begin{tabular}{ccccccc}
\toprule
Answerer & Paraphraser & Dataset & $n$ & Acc (orig) & Acc (para) & $\Delta$ (drop) \\
\midrule
\Qwen & \Mistral & ARC-Easy     & \NArcEasy  & \AccEasyOrig  & \AccEasyPara  & \DropEasy  \\
\Qwen & \Mistral & ARC-Challenge& \NArcChal  & \AccChalOrig  & \AccChalPara  & \DropChal  \\
\Mistral & \Qwen & ARC-Easy     & \NArcEasy  & \AccEasyOrigSec  & \AccEasyParaSec  & \DropEasySec  \\
\Mistral & \Qwen & ARC-Challenge& \NArcChal  & \AccChalOrigSec  & \AccChalParaSec  & \DropChalSec  \\
\bottomrule
\end{tabular}%
}
\label{tab:main}
\end{table}

%========================

\section{Discussion}

\paragraph{Contamination vs.\ robustness.}
The observed accuracy drop under paraphrase indicates that models may be relying on brittle
surface--form patterns rather than robust semantic generalization.
This raises the question of contamination:
if benchmark items or near duplicates appeared in pretraining corpora,
high performance on the original phrasing may reflect memorization rather than reasoning.
Paraphrasing disrupts such surface matches, exposing whether the model has
internalized underlying concepts or simply memorized familiar strings.
While our study does not perform explicit contamination auditing,
methods such as $n$-gram overlap checks or retrieval-based similarity~\citep{dong_generalization_2024}
could strengthen causal attribution in future work.

\paragraph{Instruction tuning sensitivity.}
Instruction tuning strongly conditions both the answerer and the paraphraser.
On the answering side, small prompt variations can flip predictions,
suggesting that instruction templates may inadvertently favor one phrasing style over another.
On the paraphrasing side, instruction tuning interacts with the model’s generative priors,
sometimes yielding paraphrases that are formally valid but less faithful semantically.
Thus, prompt design and instruction alignment are not neutral components of the pipeline,
but active factors shaping robustness outcomes.

\paragraph{Paraphrase fidelity.}
Despite targeted prompts and automated cleaning, occasional semantic drift occurs in paraphrases.
Manual inspection revealed cases where models introduced or omitted information,
which neither cleaning heuristics nor retries could fully correct.
Because the framework relies on an LLM to generate paraphrases,
paraphrase quality becomes a limiting factor for evaluation fidelity.
Improving this component---for example by using human-in-the-loop filtering,
specialized paraphrasing models, or multi-pass verification---would increase confidence
that measured drops stem from answerer brittleness rather than paraphrase artifacts.

\paragraph{Joint influence of answerer and paraphraser.}
By design, the paraphraser is always from a different model family than the answerer,
avoiding leakage through stylistic self-imitation.
However, this also means that results reflect the \emph{interaction} of two models,
not the answerer in isolation.
If the paraphraser generates awkward or biased rewrites,
measured brittleness may partially reflect its limitations.
Future protocols could disentangle these roles more cleanly,
for example by evaluating answerers against a fixed, high-fidelity paraphrase set.

\paragraph{Format dependence.}
Forcing a letter-only output format simplifies scoring but also interacts with instruction tuning.
Some errors appear to arise from rigid formatting constraints,
rather than from the model’s underlying knowledge.
Although our strict parser reduces spurious matches,
output-format sensitivity highlights that evaluation design decisions
can influence reported robustness.

\paragraph{Scope and generalization.}
Our study is limited to the ARC benchmark, which targets grade-school science.
While useful for controlled analysis, these results may not directly transfer to broader tasks such as open-domain QA, multi-turn dialogue, or reasoning-intensive benchmarks. Extending paraphrase stress tests across domains and task formats
is an important direction for future work.

%========================
\section{Conclusion}

We evaluated two 7B instruction models (\Mistral, \Qwen) under a paraphrase stress test
and found consistent accuracy drops of $6$--$10$ points on ARC--Easy and ARC--Challenge.
This indicates that strong benchmark scores may partly reflect memorization
or reliance on brittle surface patterns rather than robust reasoning.
The results underscore the importance of paraphrase-aware evaluation
and point to future work on higher-fidelity paraphrasing, contamination auditing,
and extending stress tests beyond ARC to broader tasks and domains.

%========================
%\section{Reproducibility}
%We release the code (prompts/decoding/quantization) and model/dataset identifiers.

% ACKNOWLEDGEMENTS =====
\begin{ack}

Thanks to the open-source community for releasing models, datasets, and tools that make public-only evaluations feasible.

\end{ack}

% REFERENCES ===========
\bibliographystyle{plainnat} % or unsrtnat, abbrvnat, etc.
\bibliography{references}

\begin{thebibliography}{11}
\providecommand{\natexlab}[1]{#1}
\providecommand{\url}[1]{\texttt{#1}}
\expandafter\ifx\csname urlstyle\endcsname\relax
  \providecommand{\doi}[1]{doi: #1}\else
  \providecommand{\doi}{doi: \begingroup \urlstyle{rm}\Url}\fi

\bibitem[{Allen AI}(2025)]{allen_ai_allenaiai2_arc_2025}
{Allen AI}.
\newblock allenai/ai2\_arc ({Hugging} {Face} {Datasets}), 2025.
\newblock URL \url{https://huggingface.co/datasets/allenai/ai2_arc}.

\bibitem[Carlini et~al.(2021)Carlini, Tramer, Wallace, Jagielski, Herbert-Voss, Lee, Roberts, Brown, Song, Erlingsson, Oprea, and Raffel]{carlini_extracting_2021}
Nicholas Carlini, Florian Tramer, Eric Wallace, Matthew Jagielski, Ariel Herbert-Voss, Katherine Lee, Adam Roberts, Tom Brown, Dawn Song, Ulfar Erlingsson, Alina Oprea, and Colin Raffel.
\newblock Extracting {Training} {Data} from {Large} {Language} {Models}, June 2021.
\newblock URL \url{http://arxiv.org/abs/2012.07805}.
\newblock arXiv:2012.07805 [cs].

\bibitem[Clark et~al.(2018)Clark, Cowhey, Etzioni, Khot, Sabharwal, Schoenick, and Tafjord]{clark_think_2018}
Peter Clark, Isaac Cowhey, Oren Etzioni, Tushar Khot, Ashish Sabharwal, Carissa Schoenick, and Oyvind Tafjord.
\newblock Think you have {Solved} {Question} {Answering}? {Try} {ARC}, the {AI2} {Reasoning} {Challenge}, March 2018.
\newblock URL \url{http://arxiv.org/abs/1803.05457}.
\newblock arXiv:1803.05457 [cs].

\bibitem[Dong et~al.(2024)Dong, Jiang, Liu, Jin, Gu, Yang, and Li]{dong_generalization_2024}
Yihong Dong, Xue Jiang, Huanyu Liu, Zhi Jin, Bin Gu, Mengfei Yang, and Ge~Li.
\newblock Generalization or {Memorization}: {Data} {Contamination} and {Trustworthy} {Evaluation} for {Large} {Language} {Models}, May 2024.
\newblock URL \url{http://arxiv.org/abs/2402.15938}.
\newblock arXiv:2402.15938 [cs].

\bibitem[Jiang et~al.(2023)Jiang, Sablayrolles, Mensch, Bamford, Chaplot, Casas, Bressand, Lengyel, Lample, Saulnier, Lavaud, Lachaux, Stock, Scao, Lavril, Wang, Lacroix, and Sayed]{jiang_mistral_2023}
Albert~Q. Jiang, Alexandre Sablayrolles, Arthur Mensch, Chris Bamford, Devendra~Singh Chaplot, Diego de~las Casas, Florian Bressand, Gianna Lengyel, Guillaume Lample, Lucile Saulnier, Lélio~Renard Lavaud, Marie-Anne Lachaux, Pierre Stock, Teven~Le Scao, Thibaut Lavril, Thomas Wang, Timothée Lacroix, and William~El Sayed.
\newblock Mistral {7B}, October 2023.
\newblock URL \url{http://arxiv.org/abs/2310.06825}.
\newblock arXiv:2310.06825 [cs].

\bibitem[Kandpal et~al.(2022)Kandpal, Wallace, and Raffel]{kandpal_deduplicating_2022}
Nikhil Kandpal, Eric Wallace, and Colin Raffel.
\newblock Deduplicating {Training} {Data} {Mitigates} {Privacy} {Risks} in {Language} {Models}, December 2022.
\newblock URL \url{http://arxiv.org/abs/2202.06539}.
\newblock arXiv:2202.06539 [cs].

\bibitem[Liu et~al.(2023)Liu, Xia, Wang, and Zhang]{liu_is_2023}
Jiawei Liu, Chunqiu~Steven Xia, Yuyao Wang, and Lingming Zhang.
\newblock Is {Your} {Code} {Generated} by {ChatGPT} {Really} {Correct}? {Rigorous} {Evaluation} of {Large} {Language} {Models} for {Code} {Generation}, October 2023.
\newblock URL \url{http://arxiv.org/abs/2305.01210}.
\newblock arXiv:2305.01210 [cs].

\bibitem[{Mistral AI}(2025)]{mistral_ai_mistralaimistral-7b-instruct-v03_2025}
{Mistral AI}.
\newblock mistralai/{Mistral}-{7B}-{Instruct}-v0.3 ({Model} {Card}), 2025.
\newblock URL \url{https://huggingface.co/mistralai/Mistral-7B-Instruct-v0.3}.

\bibitem[{Qwen Team}(2025)]{qwen_team_qwenqwen25-7b-instruct_2025}
{Qwen Team}.
\newblock Qwen/{Qwen2}.5-{7B}-{Instruct} ({Model} {Card}), 2025.
\newblock URL \url{https://huggingface.co/Qwen/Qwen2.5-7B-Instruct}.

\bibitem[Ribeiro et~al.(2020)Ribeiro, Wu, Guestrin, and Singh]{ribeiro_beyond_2020}
Marco~Tulio Ribeiro, Tongshuang Wu, Carlos Guestrin, and Sameer Singh.
\newblock Beyond {Accuracy}: {Behavioral} {Testing} of {NLP} models with {CheckList}, May 2020.
\newblock URL \url{http://arxiv.org/abs/2005.04118}.
\newblock arXiv:2005.04118 [cs].

\bibitem[Yang et~al.(2024)Yang, Yang, Hui, Zheng, Yu, Zhou, Li, Li, Liu, Huang, Dong, Wei, Lin, Tang, Wang, Yang, Tu, Zhang, Ma, Yang, Xu, Zhou, Bai, He, Lin, Dang, Lu, Chen, Yang, Li, Xue, Ni, Zhang, Wang, Peng, Men, Gao, Lin, Wang, Bai, Tan, Zhu, Li, Liu, Ge, Deng, Zhou, Ren, Zhang, Wei, Ren, Liu, Fan, Yao, Zhang, Wan, Chu, Liu, Cui, Zhang, Guo, and Fan]{yang_qwen2_2024}
An~Yang, Baosong Yang, Binyuan Hui, Bo~Zheng, Bowen Yu, Chang Zhou, Chengpeng Li, Chengyuan Li, Dayiheng Liu, Fei Huang, Guanting Dong, Haoran Wei, Huan Lin, Jialong Tang, Jialin Wang, Jian Yang, Jianhong Tu, Jianwei Zhang, Jianxin Ma, Jianxin Yang, Jin Xu, Jingren Zhou, Jinze Bai, Jinzheng He, Junyang Lin, Kai Dang, Keming Lu, Keqin Chen, Kexin Yang, Mei Li, Mingfeng Xue, Na~Ni, Pei Zhang, Peng Wang, Ru~Peng, Rui Men, Ruize Gao, Runji Lin, Shijie Wang, Shuai Bai, Sinan Tan, Tianhang Zhu, Tianhao Li, Tianyu Liu, Wenbin Ge, Xiaodong Deng, Xiaohuan Zhou, Xingzhang Ren, Xinyu Zhang, Xipin Wei, Xuancheng Ren, Xuejing Liu, Yang Fan, Yang Yao, Yichang Zhang, Yu~Wan, Yunfei Chu, Yuqiong Liu, Zeyu Cui, Zhenru Zhang, Zhifang Guo, and Zhihao Fan.
\newblock Qwen2 {Technical} {Report}, September 2024.
\newblock URL \url{http://arxiv.org/abs/2407.10671}.
\newblock arXiv:2407.10671 [cs].

\end{thebibliography}

\end{document}